\documentclass[fleqn,10pt]{wlpeerj}
\pdfoutput=1
\usepackage{multirow}
\usepackage[T1]{fontenc}

\title{Two-stage training algorithm for AI robot soccer}

\author[1]{Taeyoung Kim}
\author[1]{Luiz Felipe Vecchietti}
\author[1]{Kyujin Choi}
\author[2]{Sanem Sariel}
\author[1]{Dongsoo Har}
\affil[1]{Cho Chun Shik Graduate School of Green Transportation, Korea Advanced Institute of Science and Technology, Daejeon 34051, South Korea}
\affil[2]{Istanbul Technical University, Department of Computer Engineering, Istanbul, 34496, TURKEY}
\corrauthor[1]{Dongsoo Har}{dshar@kaist.ac.kr}

\begin{abstract}
In multi-agent reinforcement learning, the cooperative learning behavior of agents is very important. In the field of heterogeneous multi-agent reinforcement learning, cooperative behavior among different types of agents in a group is pursued. Learning a joint-action set during centralized training is an attractive way to obtain such cooperative behavior, however, this method brings limited learning performance with heterogeneous agents. To improve the learning performance of heterogeneous agents during centralized training, two-stage heterogeneous centralized training which allows the training of multiple roles of heterogeneous agents is proposed. During training, two training processes are conducted in a series. One of the two stages is to attempt training each agent according to its role, aiming at the maximization of individual role rewards. The other is for training the agents as a whole to make them learn cooperative behaviors while attempting to maximize shared collective rewards, e.g., team rewards. Because these two training processes are conducted in a series in every timestep, agents can learn how to maximize role rewards and team rewards simultaneously. The proposed method is applied to 5 versus 5 AI robot soccer for validation. Simulation results show that the proposed method can train the robots of the robot soccer team effectively, achieving higher role rewards and higher team rewards as compared to other approaches that can be used to solve problems of training cooperative multi-agent.
\end{abstract}

\begin{document}

\flushbottom
\maketitle
\thispagestyle{empty}

% === I. Introduction ===
\section*{Introduction}
\label{sec:1}

Recently, deep reinforcement learning (DRL) has been widely applied to deterministic games \cite{silver2018general}, video games \citep{mnih2015human, mnih2016asynchronous, silver2016mastering}, and complex robotic tasks \citep{andrychowicz2017hindsight, hwangbo2019learning, seo2019rewards, vecchietti2020batch}. Despite the breakthrough results achieved in the field of DRL, deep learning in multi-agent environments that require both cooperation and competition is still challenging. Promising results have been for cooperative-competitive multi-agent games such as StarCraft \citep{vinyals2019grandmaster} and Dota \citep{berner2019dota}. For multi-agent problems such as multi-robot soccer \citep{liu2019emergent}, security \citep{he2015improving, klima2016markov}, traffic control \citep{chu2019multi, zhang2019cityflow}, and autonomous driving \citep{shalev2016safe, sallab2017deep}, non-stationarity, partial observability, multi-agent training schemes, and heterogeneity can be challenging issues \citep{nguyen2020deep}. To solve these challenges, multi-agent reinforcement learning (MARL) techniques \citep{lowe2017multi, sunehag2017value, foerster2018counterfactual, vinyals2019grandmaster, liu2019emergent, samvelyan2019starcraft, rashid2020monotonic} have been intensively investigated.

When using the MARL, several works have used the centralized training in decentralized execution (CTDE) framework \citep{lowe2017multi, sunehag2017value, foerster2018counterfactual, rashid2020monotonic}. In the CTDE framework, local observations of agents, global state of the environment, and joint-actions taken by the agents at each timestep are available during training to the centralized policy network, while only the local observations of agents are available during execution. In other words, each agent selects its action, that is the output of a policy network, without considering the full information of the environment. To address the non-stationarity problem, multi-agent deep deterministic policy gradient (MADDPG) \citep{lowe2017multi} was proposed using a CTDE framework and the DDPG actor-critic algorithm for continuous action spaces \citep{lillicrap2015continuous}. When cooperative behavior is to be achieved, representing that there is a cooperative reward that should be maximized by multiple agents, credit should be assigned accordingly to each agent based on its contribution. To address this problem, counterfactual multi-agent (COMA) \citep{foerster2018counterfactual}, value decomposition networks (VDN) \citep{sunehag2017value}, and monotonic value function factorization (QMIX) \citep{rashid2020monotonic} have been proposed, using the CTDE framework combined with value-based algorithms such as deep Q networks (DQN) \citep{mnih2013playing}, deep recurrent Q networks (DRQN) \citep{hausknecht2015deep}, and dueling Q networks \citep{wang2016dueling}.

In this paper, a novel training method for MARL of heterogeneous agents, in which each agent should choose its action in a decentralized manner, is proposed. The proposed method addresses how to provide an optimal policy and maximize the cooperative behavior among heterogeneous agents. To this end, during training, two training stages are conducted in a series. The first stage is for making each agent learn to maximize its individual role reward while executing its individual role. The second one is for making the agents as a whole learn cooperative behavior, aiming at the maximization of team reward. The proposed method is designed to be applied to MARL with heterogeneous agents in cooperative or cooperative-competitive scenarios. In this paper, a cooperative-competitive AI robot soccer environment is used for experiments. The environment can be described in relation to 5 versus 5 robot soccer game described in \cite{hong2021}. In the robot soccer game, two teams of five robots capable of kick and jump behaviors compete against each other, similarly to the StarCraft, so the game can be seen as a micro-management problem. The policy for the proposed method and other methods for comparisons are trained by using self-play \citep{heinrich2015fictitious, lanctot2017unified, silver2017mastering}. Self-play in a competitive environment is used so that the opponent team is kept at an appropriate level of difficulty at each training stage.

The main contributions of this paper are as follows

\begin{enumerate}
\item A framework for novel training method called two-stage heterogeneous centralized training (TSHCT) aiming at centralized training of heterogeneous agents is proposed. In the proposed method, there are two training stages that are conducted in a series. The first stage is responsible for training individual behaviors by maximizing individual role rewards. The second stage is for training cooperative behaviors by maximizing a shared collective reward.
\item Experiments are conducted to compare the performance of the proposed method with those of other methods. The proposed method and the baseline methods are trained with self-play. To compare the results obtained from the experiments, total rewards (during training) and score/concede rates (against different opponent teams) are presented. From the comparisons, we will show better performance of the proposed method during game
\item The proposed method aims at MARL with heterogeneous agents in cooperative and cooperative-competitive scenarios. For experiments, a cooperative-competitive AI robot soccer environment, where there are 5 robots with 3 different roles in each team (one goalkeeper, two defenders, and two forwards), is used.
\end{enumerate}

The remainder of this paper is organized as follows. Section 2 presents the concept of the MARL, system modeling, and other methods which are used as baselines for comparisons in the experiments. Section 3 introduces the proposed method in details. Section 4 presents the simulation environment, ablation studies, and game results of the AI robot soccer. Section 5 concludes this paper.

% === II. Background ===
\section*{Background}
\label{sec:2}

In this section, the system modeling relevant to the proposed method is presented. Also, other methods for cooperative MARL using the CTDE framework are presented.

\subsection*{System Modeling}

The cooperative-competitive multi-agent problem, specifically applied in this paper to AI robot soccer, is modeled as a decentralized partially observable Markov decision process (Dec-POMDP) \citep{oliehoek2016concise} that each agent has its own observation of the environment. The Dec-POMDP can be formulated by a tuple $G = <S,U,P,r,Z,O,n,\gamma>$. The set of states and the set of actions are represented by $S$ and $U$ respectively. Each team contains $n$ agents. The observation function $O(s,a)$ determines the observation $z \in Z$ that each agent perceived individually at each time step. At each time step, the $n$ agents choose their actions $u^a \in U, a=1,...,n,$ based on their action-observation history. In this modeling, as recurrent neural networks (RNN) \citep{hochreiter1997long} is used by the MARL algorithm, the policy is conditioned on the joint action-observation history as well as the current agent observation $z$. The state of the environment changes according to a transition probability $P$. Unlike the partially observable stochastic game, all agents in Dec-POMDP share a collective reward and an individual reward drawn from the reward function $r(s,\mathbf{u})$. The discount factor of the MARL algorithm is represented by $\gamma$.

In MARL, as multiple agents act simultaneously in the environment based only on their own action-observation history and do not know about the individual policy of each agent, there exists a non-stationarity problem. The behaviors of other agents are changing during training and can influence the reward received by each agent. To address this issue, the system is modeled using a centralized training in decentralized execution (CTDE) framework. In the CTDE framework, the full state of the environment can be accessed in the training procedure to get the state-action value. On the other hand, only the local observation can be accessed by the agent during execution. The joint-action from all agents is also available during the training procedure by the centralized policy to alleviate the non-stationarity issue.

In this paper, we focus on value-based MARL algorithms applied in environments where a sense of cooperation is needed between agents, meaning that they share a collective reward. The proposed algorithm is to be combined with deep recurrent Q-networks (DRQN) \citep{hausknecht2015deep} and dueling deep Q-networks \citep{wang2016dueling}. The DRQN algorithm, as proposed in \cite{hausknecht2015deep}, addresses single-agent with partially observable environments. The architecture consists of the DQN \citep{mnih2015human} combined with RNN. The DRQN approximates the state-action value function $Q(o,u)$ with RNN to maintain an internal state and aggregate observations over time. It also can be taken to approximate $Q(o_t,h_{t-1},u)$, where $h_{t-1}$ represents the hidden state, which has information of previous states and acts as a memory. The proposed method is also to be combined with the dueling deep Q-networks \citep{wang2016dueling}. The dueling deep Q-networks is a neural network architecture designed for value-based RL that has two streams in the computation of the state-action value. One stream is for approximating the value function $V(s)$ and the other is for approximating the advantage function $A(s,u)$. The value function $V(s)$ depends only on state and presents how good a state is. The advantage function $A(s,u)$ depends on both state and action and presents how advantageous it is to take an action $u$ in comparison to the other actions at the given state $s$. The value and the advantage are merged to get the final state-action value $Q(s,u)$ as follows
\begin{equation}
    Q(s,u) = V(s) + A(s,u) - {\frac{\sum_{u'} A(s,u')}{N}},
    \label{eq:dueling}
\end{equation}
where $u'$ represents each possible action and N is the number of actions. In this paper, the dueling deep Q-networks is combined with the RNN to handle the action-observation history used as the input of the policy. In the architecture of dueling deep Q-networks with the RNN, e.g., Dueling DRQN, the RNN is inserted right before the crossroad of streams of computation. The dueling DRQN is compared with the DRQN as an ablation study in Section 4.

% Q(s,u) = V(s) + A(s,u) - {{\sum_{u'} A(s,u')} \over {N}}

In the following subsections, other methods relevant to comparisons are presented. In this paper, we focus on methods that can be combined with off-policy value-based algorithms and focus on the maximization of a joint state-action value, trying to assign proper credit to individual agents on the shared reward received.

\subsection*{Counterfactual multi-agent Policy Gradients}

Counterfactual multi-agent (COMA), introduced by \cite{foerster2018counterfactual}, utilizes a single centralized critic to train decentralized actors and deals with the challenge of the multi-agent credit assignment problem. In the cooperative environments that are the main target for the COMA, it is difficult to determine the contribution of each agent to the shared collective reward received by the team. The centralized critic has access to the global state and the actions of the agent to model the joint state-action value function.

\subsection*{Value Decomposition Network}

The value decomposition network (VDN) \citep{sunehag2017value} aims at learning a joint-action value function $Q_{tot}(\tau,u)$, where $\tau$ is a joint-action observation history and $u$ is a joint-action. The $Q_{tot}(\tau,u)$ can be expressed as a sum of individual value functions $Q_{i}(\tau^{i},u^{i};\theta^{i})$ as follows
\begin{equation}
    Q_{tot}(\tau,u) = \sum_{i=1}^{n} Q_{i}(\tau^{i},u^{i};\theta^{i}),
    \label{eq:vdn}
\end{equation}
where each $Q_{i}(\tau^i,u^i;\theta^i)$ is a utility function of the $i$-th agent and $\theta^{i}$ is the policy of the $i$-th agent. The loss function for the VDN is the same as that of the deep Q-network (DQN) \citep{mnih2015human}, where $Q$ is replaced by $Q_{tot}(\tau,u)$.

\subsection*{QMIX}

QMIX \citep{rashid2020monotonic} is a deep multi-agent reinforcement learning method to be trained using CTDE. It uses the additional global state information that is the input of a mixing network. The QMIX is trained to minimize the loss, just like the VDN \citep{sunehag2017value}, given as
\begin{equation}
    \mathcal{L}(\theta) = \sum_{i=1}^{b} [(y^{tot}_{i} - Q_{tot}(\tau,u,s;\theta))^2],
    \label{eq:qmix_loss}
\end{equation}
where b is the batch size of transitions sampled from the replay buffer and $Q_{tot}$ is output of the mixing network and the target $y^{tot}_i = r + \gamma max_{u'} Q_{tot}(\tau',u',s';\theta^{-})$, and $\theta^{-}$ are the parameters of a target network. The QMIX allows learning of joint-action-value functions, which are equivalent to the composition of optimal Q values of each agent. This is achieved by imposing a monotonicity constraint on the mixing network. Monotonicity can be enforced by the constraint on the relationship between $Q_{tot}$ and individual $Q$ value functions, given as
\begin{equation}
    Q_{a} : {{\partial Q_{tot}} \over {\partial Q_{a}}} \geq 0, \forall a \in A.
    \label{eq:qmix}
\end{equation}

% === III. Proposed Method ===
\section*{Proposed Method}
\label{sec:3}

In heterogeneous multi-agent reinforcement learning, the main challenge can be described as how to provide an optimal policy and maximize cooperative behavior in a heterogeneous multi-agent environment. In this scenario, the agents act independently and maximize not only the individual reward but also a shared reward. To tackle this problem, a novel training method called two-stage heterogeneous centralized training is proposed and described in this section and to be applied to 5 versus 5 AI robot soccer.

\subsection*{MARL Structure for AI Robot Soccer}

\begin{figure}[h!]
\centering
\includegraphics[width=\linewidth]{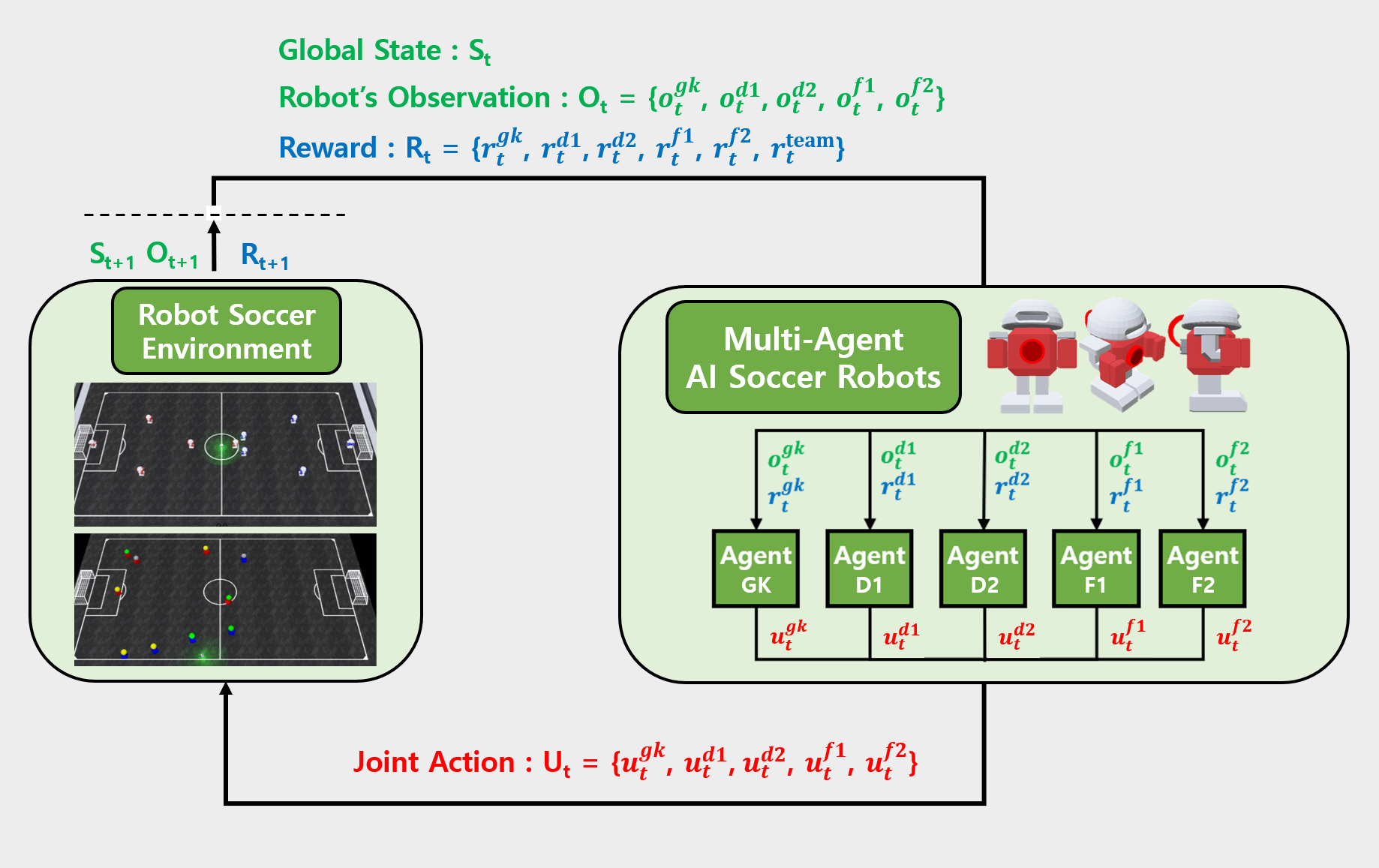}
\caption{MARL structure for AI robot soccer.}
\label{fig:structure}
\end{figure}

The MARL structure in 5 versus 5 AI robot soccer is presented in Fig.~\ref{fig:structure}. Each robot has individual observations and individual rewards according to its role in soccer game, e.g. goalkeeper, defender, or forward. Each robot receives its individual observation at each timestep and selects its action according to a policy network that also takes into consideration past individual observations and actions taken. The concatenation of individual actions of the 5 robots forms a joint-action set. By performing this joint-action in the AI robot soccer environment, the global state, individual observations, and individual role rewards are calculated and communicated with the agents. It is noted that the global state is available only during training.

\subsection*{TSHCT Architecture}

\begin{figure}[h!]
\centering
\includegraphics[width=\linewidth]{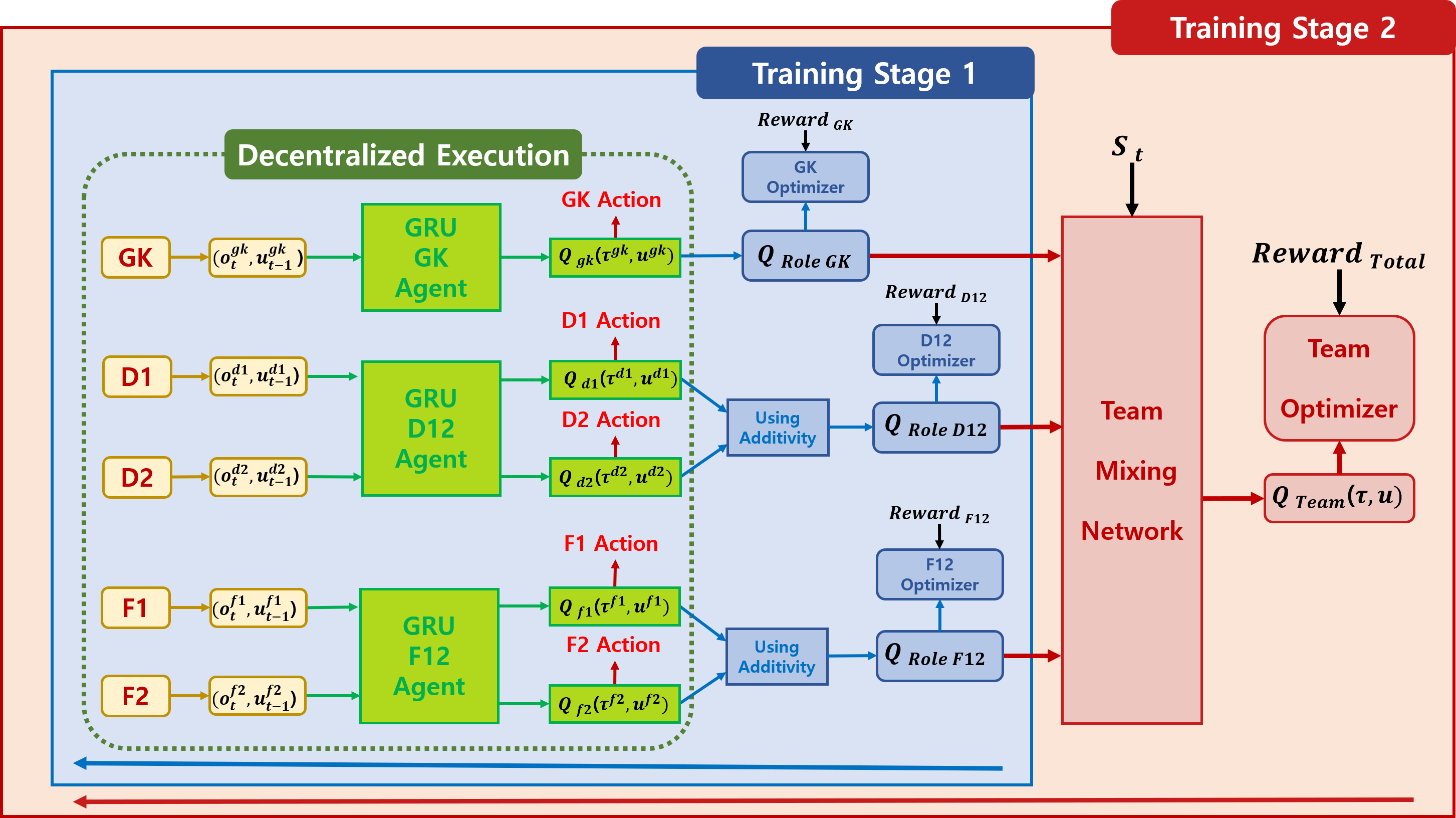}
\caption{Overall architecture of two-stage heterogeneous centralized training. In the first stage, the agents are trained using their individual role rewards, a goalkeeper reward, a defender reward, and an attacker reward. In the second stage, the agents are trained using a collective team reward, represented by total reward. The global state of the environment $s_t$ is also used as an input of the team mixing network. A shared policy is used by defenders and by forwards.}
\label{fig:architecture}
\end{figure}

As shown in Fig.~\ref{fig:architecture}, the training procedure is divided into two stages. In the first stage, agents of the same type (homogeneous agents, e.g., 2 agents as defenders) are trained. Decentralized execution is used during inference and a shared policy is used by the agents of the same type. In the second training stage, all heterogeneous agents are trained jointly. These two stages are executed in a serial learning structure.

To model each agent's policy, the structure of DQN with gated recurrent unit (GRU) \citep{chung2014empirical} or the structure of Dueling Q-Networks with GRU is used in the experiments. The policy network receives as input 40 subsequential frames with the current individual observation of the agent $o_{t}^{(N_n)}$ and the last action chosen $u_{(t-1)}^{(N_n)}$, where $N_n$ is the $n$-th agent of the $N$-role (type). The output of the policy network is the state-action value $Q_{N_n}$. The action with the highest Q-value is chosen at each timestep in a greedy fashion.

In training stage 1, the $Q_{(Role N)} \forall N \in (GoalKeeper, Defenders, Forwards))$ is calculated by adding Q-values $Q_{N_n}$ from the homogeneous agent network. In training stage 2, the team mixing network combines the individual role rewards into the shared collective reward. The mixing network is modeled as a hypernetwork \citep{ha2016hypernetworks}, using feed-forward layers. The hypernetwork is conditioned on the global state $S_t$ of the environment and takes the values of $Q_{(Role GK)}$, $Q_{(Role D12)}$, and $Q_{(Role F12)}$ produced in training stage 1 as inputs. The output of the mixing network is $Q_{Team}$.

\subsection*{TSHCT Learning Equations}

The proposed method is used to minimize the losses through the entire training. In training stage 1, each role optimizer updates the weights of the policy network to minimize the loss $\mathcal{L}_{Role N}(\theta)$ in relation to the target $y^{Role N}$ calculated with individual role rewards $Reward_{Role N}$ based on the Bellman equation and given as follows
\begin{equation*}
    y^{Role N} = Reward_{Role N} + \gamma max_{u'} Q_{Role N} (\tau',u',s';\theta^{-}),
\end{equation*}
\begin{equation}
    \mathcal{L}_{Role N}(\theta) = \sum_{i=1}^{b} [(y_i^{Role N} - Q_{Role N}(\tau,u,s;\theta))^2],
    \label{eq:role_loss}
\end{equation}
where $\gamma$ and $\theta^{-}$ are the parameters of a target network, the discount factor and policy, similar to the ones presented in DQN \citep{mnih2015human} to stabilize the training procedure and $b$ is the batch size of episodes sampled from the replay buffer.
In training stage 2, the team optimizer updates the weights of mixing network and policy networks to minimize the team loss in relation to the team target $y^{Team}$ calculated with the total shared reward $Reward_{Total}$, which is the sum of sparse cooperative team rewards and dense individual role rewards. The team loss $\mathcal{L}_{Team}(\theta)$ is given as follows
\begin{equation*}
    y^{Team} = Reward_{Total} + \gamma max_{u'} Q_{Team}(\tau',u',s';\theta^{-}),    
\end{equation*}
\begin{equation}
    \mathcal{L}_{Team}(\theta) = \sum_{i=1}^{b} [(y_i^{Team} - Q_{Team}(\tau,u,s;\theta))^2].
    \label{eq:team_loss}
\end{equation}

Equations~\ref{eq:role_loss} and ~\ref{eq:team_loss} are analogous to the minimum squared loss used in \cite{mnih2015human}. Using additivity \citep{sunehag2017value} and monotonicity \citep{rashid2020monotonic}, the TSHCT trains heterogeneous agents by maximizing $Q_{Team}$ in stage 2, while learning multiple roles by maximizing the Q-value of each individual role $Q_{(Role GK)}$, $Q_{(Role D12)}$, and $Q_{(Role F12)}$ in stage 1.

\subsection*{TSHCT Curriculum Learning through Self-Play}

To train a robust policy in a competitive-cooperative scenario that can work well against multi-agent in the opponent team, curriculum learning is needed. In this paper, we use self-play as a form of the implicit curriculum with the objective of learning robust AI robot soccer strategies. The implicit self-play curriculum is implemented by updating the opponent team when the number of episodes reaches a particular number. The opponent team is kept updated and reference policies take turns. Using self-play, it is possible to keep the opponent team at an appropriate level of competitivity, not too strong so that the policy allows good behavior and not too easy so that the policy avoids weak behaviors. The soccer strategy learned through self-play tends to lead to acceptable game performance \citep{heinrich2015fictitious, lanctot2017unified, silver2017mastering} as the result of the automated curriculum.

% === IV. Simulation Results ===
\section*{Simulation Results}
\label{sec:4}

In this section, the MARL environment used in the experiments and the results obtained by the TSHCT and other baseline methods are described.

\subsection*{AI Robot Soccer MARL Environment}

\begin{figure}[h!]
\centering
\includegraphics[width=\linewidth]{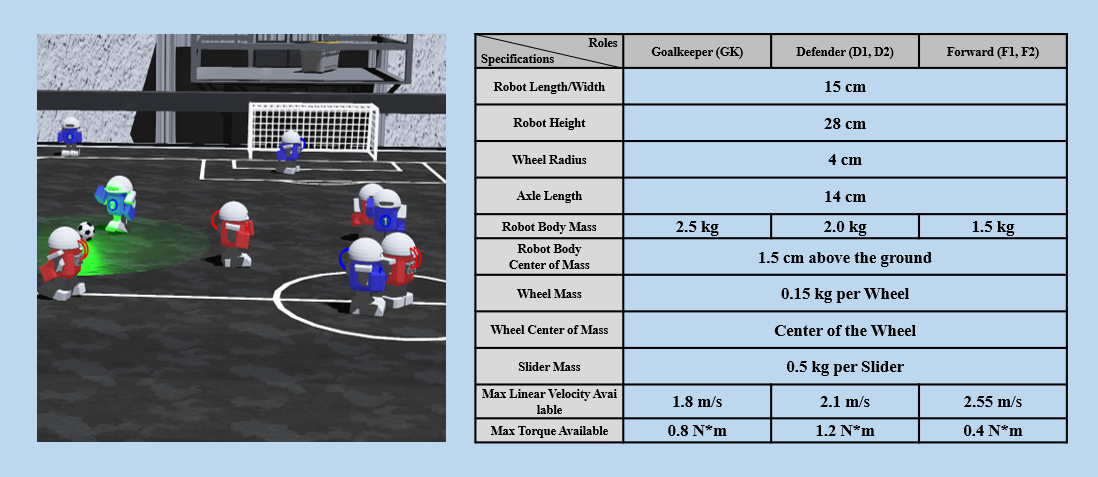}
\caption{Specifications of the AI robot soccer environment. Robots with different roles, goalkeeper, defender, or forward, have different mass, maximum linear velocity, and maximum torque.}
\label{fig:environment}
\end{figure}

To demonstrate the performance of the TSHCT, experiments are conducted in an AI robot soccer environment with specifications presented in Fig.~\ref{fig:environment}, which is developed with Webots robot simulation software \citep{michel2004cyberbotics} and based on the environment described in \cite{hong2021}. In this AI Soccer simulation game, two teams compete similarly to a real soccer game, trying to kick the ball into the opponent’s goal area to score and to win the game against the opponent team. In each team, there are 5 robots with 3 different roles (one goalkeeper, two defenders, and two forwards). The AI robot soccer game is divided into two 5 minute-long halves. For training, the game is divided up into episodes of 40 sequential frames. An episode is over whenever 40 sequential frames are processed.

\subsubsection*{Global State and Observations}
The global state, available only during centralized training and used as input to the mixing network, contains information of all the soccer robots and the ball. Specifically, the state vector contains the coordinates and orientations of all soccer robots, including robots of the opponent team, and the ball coordinates. The coordinates are relative to the center of the field (origin). The individual local observations of each robot are their relative positions in the field and relative distances and orientations to other robots and to the ball within their range of view. These observations are used as inputs of the policy networks.

\subsubsection*{Action}
The basic actions committed by the robots are move, jump and kick. They are achieved by giving continuous control variables to the feet and legs. To achieve these behaviors a discrete set of 20 actions is designed which is allowed to be taken by the agent at each timestep. A discrete set of actions is used so that the DRQN and the Dueling DRQN can be used as the off-policy value-based algorithms for the experiments. The discrete action set consists of actions of forward motion, backward motion, 6 directions of forward turns, 4 directions of backward turns, clockwise and counterclockwise turns, 2 kinds of forward turn combined with kick, 2 kinds of forward motion combined with kick, stop combined with kick, and stop.

\subsubsection*{Reward}
To train AI soccer robots to perform their roles and cooperative behavior, individual role rewards and a shared team reward are defined. Individual role rewards are a combination of dense rewards associated with two pieces of information. One is the ball information relative to the robot, such as distance, velocity, and angle. The other is the information of the expected position which is defined for each role, i.e. default position where the robot should be to play its role. The team reward is a combination of a sparse reward related to scoring and dense rewards related to the distance and velocity between the ball and the opponent's goal.

\subsection*{Network Hyperparameters}

The neural network hyperparameters used in the experiments are as follows
\begin{itemize}
    \item DRQN architecture: 2 layers with 256 hidden units, 1 layer of GRU with 128 hidden units, and ReLU non-linearities.
    \item Dueling DRQN architecture: 1 layer with 256 hidden units, 1 layer of GRU with 128 hidden units, 1 layer with 128 hidden units for value prediction, 1 layer with 128 hidden units for advantage prediction, and ReLU non-linearities.
    \item Mixing network architecture: 1 layer of mixing network with 32 hidden units, 2 layers of hypernetworks with 32 hidden units, and ReLU non-linearities.
    \item ADAM optimizer \citep{kingma2014adam} with learning rate set to $4 \times 10^{-5}$ for both policy and mixing networks.
    \item Discount factor $\gamma$ set to 0.99.
    \item Target networks updated every 16000 iterations.
    \item Epsilon used for exploration decreased by 0.025 every $10^4$ iterations until it is kept at 0.05 at the end of training.
    \item Buffer size set to store $5 \times 10^3$ episodes.
    \item Batch size set to 64.
\end{itemize}

\subsection*{Results}

\subsubsection*{Evaluation against baselines: COMA, VDN, and QMIX}

In this section, the evaluation of the TSHCT against baseline methods, COMA, VDN, and QMIX, are presented. The proposed method and baseline methods are trained for a total of 200k episodes using epsilon greedy exploration with self-play. As the result of the evaluation, comparisons of rewards and score-concede rates are given. The score-concede rate is defined as the percentage of the number of scores divided by the sum of the number of scoring and conceding.

\begin{figure}[h!]
\centering
\includegraphics[width=0.85\linewidth]{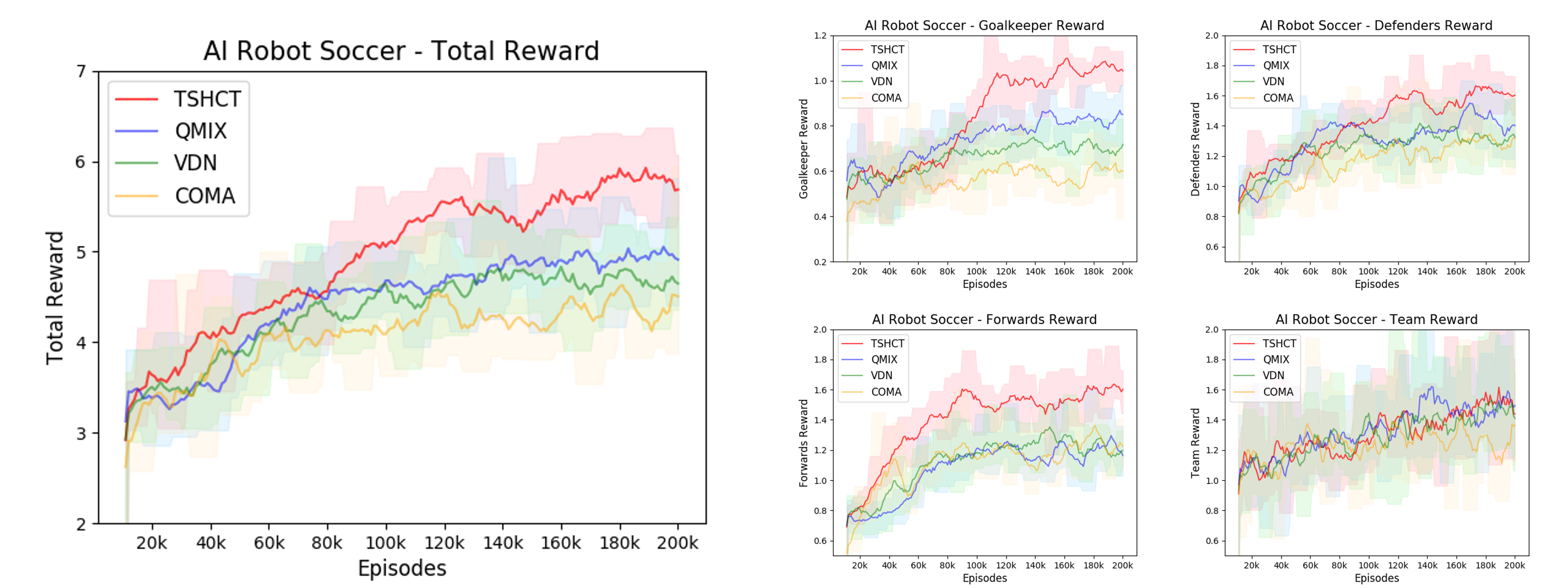}
\caption{Total reward obtained during training by TSHCT, QMIX, VDN, and COMA. It is evaluated against an opponent team trained with COMA for 200k episodes.}
\label{fig:rewards_vs_coma}
\end{figure}

\begin{figure}[h!]
\centering
\includegraphics[width=0.85\linewidth]{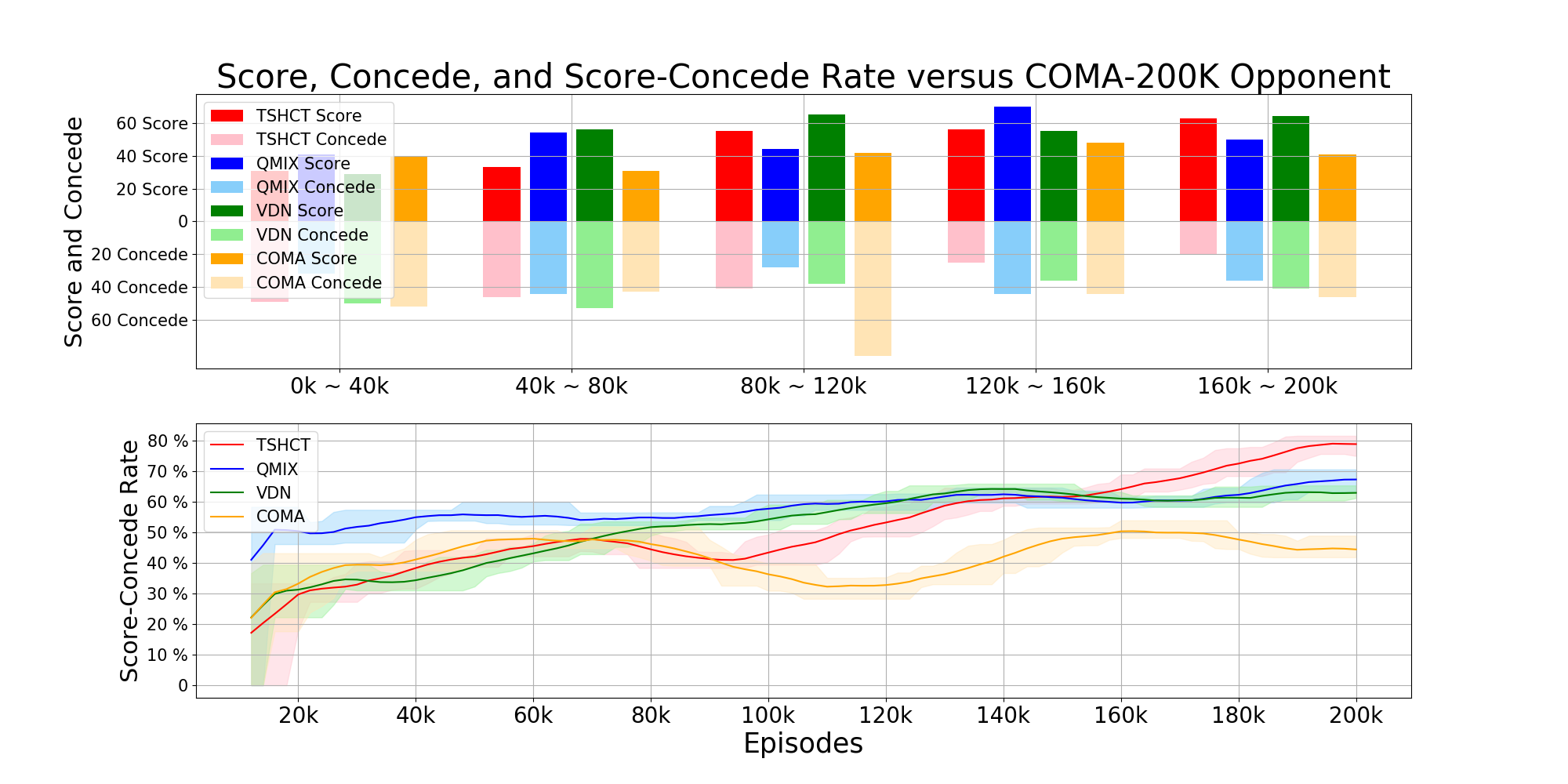}
\caption{Comparison of score, concede, and score-concede rate obtained during training by TSHCT, QMIX, VDN, and COMA. It is evaluated against an opponent team trained by COMA with 200k episodes.}
\label{fig:rate_vs_coma}
\end{figure}

In the first place for evaluation of the performance of the TSHCT, the opponent team trained by the COMA with 200k episodes is used. The experimental result shows that the TSHCT is superior to COMA, VDN, and QMIX algorithms after 80k episodes, as shown in Fig.~\ref{fig:rewards_vs_coma}. The maximum average total rewards of TSHCT, QMIX, VDN, and COMA are 5.92, 5.05, 4.83, and 4.63, respectively. The score-concede rate is defined as the maximum value of the averages of score-concede rates obtained over 10 sequential games. The score-concede rates of TSHCT, QMIX, VDN, and COMA are 79.01\%, 67.30\%, 64.21\%, and 50.40\%, respectively, as shown in Fig.~\ref{fig:rate_vs_coma}. It is observed that the TSHCT improves the score-concede rate by 28.97\% as compared to that of COMA.

\begin{figure}[h!]
\centering
\includegraphics[width=0.85\linewidth]{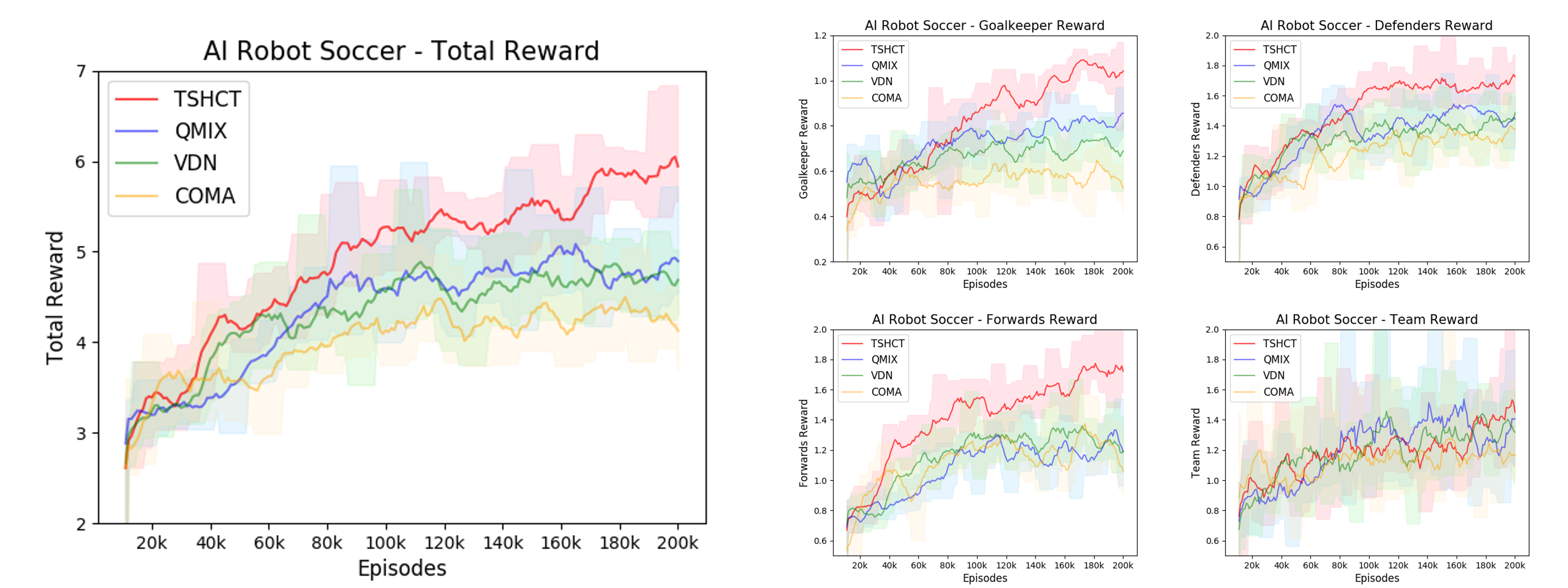}
\caption{Total reward obtained during training by TSHCT, QMIX, VDN, and COMA. It is evaluated against an opponent team trained by VDN with 200k episodes.}
\label{fig:rewards_vs_vdn}
\end{figure}

\begin{figure}[h!]
\centering
\includegraphics[width=0.85\linewidth]{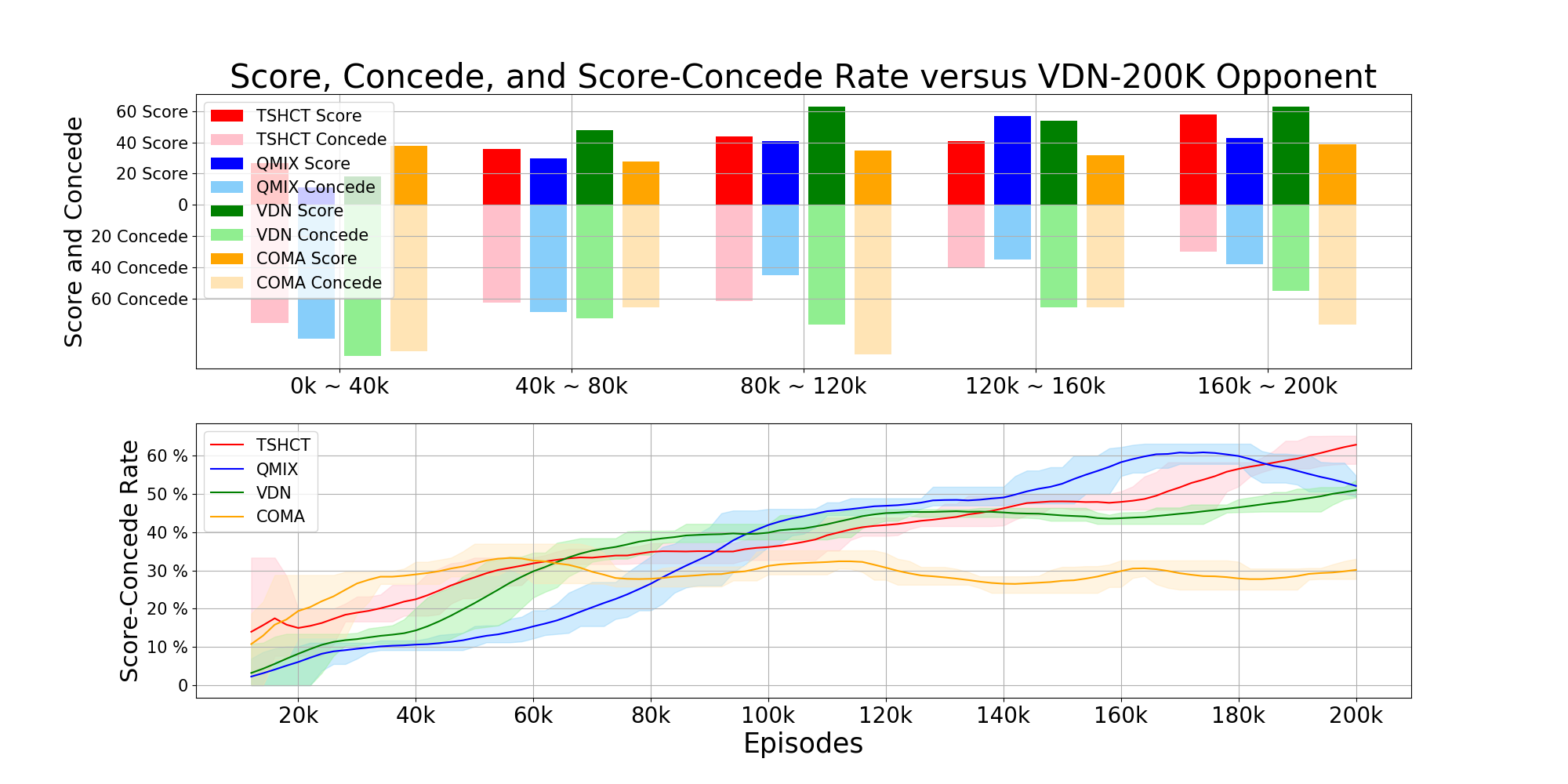}
\caption{Comparisons of score, concede, and score-concede rate obtained during training by TSHCT, QMIX, VDN, and COMA. The score, concede, and score-concede rate are evaluated against an opponent team trained by VDN with 200k episodes.}
\label{fig:rate_vs_vdn}
\end{figure}

For the other evaluation of the performance of the TSHCT, the opponent team trained by the VDN model is used. Experiment results presented in Fig.~\ref{fig:rewards_vs_vdn} show that the TSHCT is superior to the baseline algorithms after 80k episodes. The maximum average total rewards of TSHCT, QMIX, VDN, and COMA are 6.05, 5.08, 4.89, and 4.50, respectively.
The maximum averages of score-concede rate of TSHCT, QMIX, VDN, and COMA are 62.85\%, 60.85\%, 50.97\%, and 32.27\%, respectively, as shown in Fig.~\ref{fig:rate_vs_coma}. It is observed that the TSHCT improved the score-concede rate by 11.88\% as compared to that of VDN.

\begin{figure}[h!]
\centering
\includegraphics[width=0.85\linewidth]{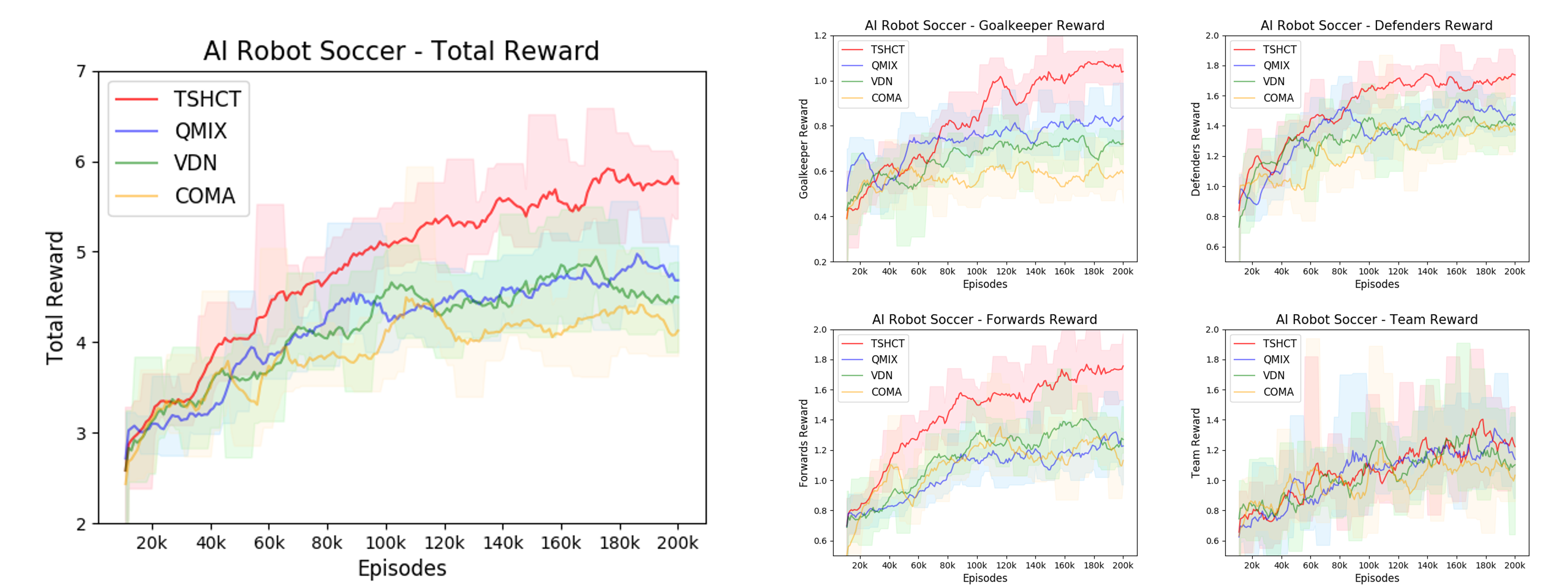}
\caption{Total reward obtained during training by TSHCT, QMIX, VDN, and COMA evaluated against an opponent team trained by QMIX with 200k episodes.}
\label{fig:rewards_vs_qmix}
\end{figure}

\begin{figure}[h!]
\centering
\includegraphics[width=0.85\linewidth]{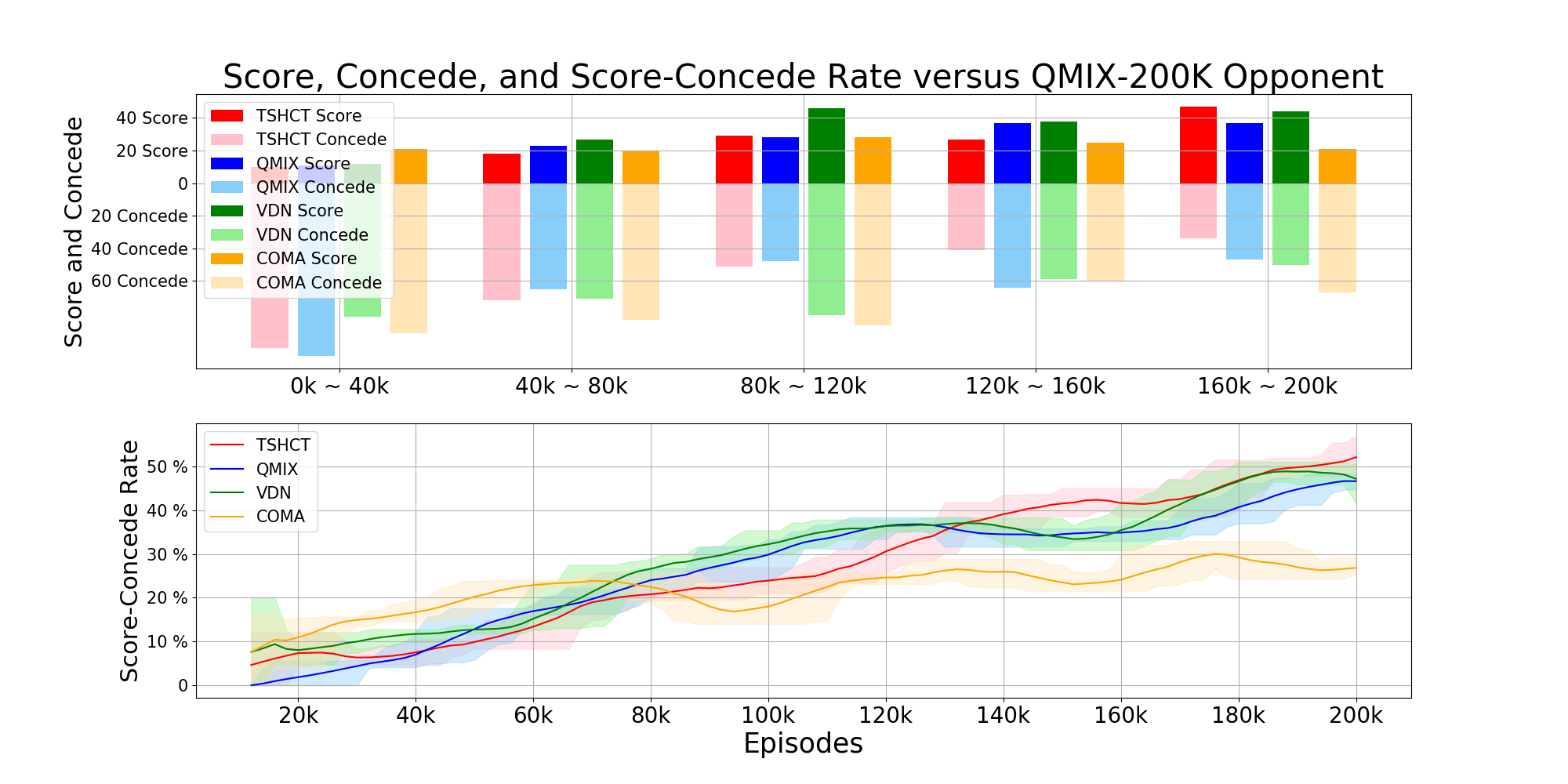}
\caption{Comparison of score, concede, and score-concede rate obtained during training by TSHCT, QMIX, VDN, and COMA. The score, concede, and score-concede rate are evaluated against an opponent team trained by QMIX with 200k episodes.}
\label{fig:rate_vs_qmix}
\end{figure}

For another evaluation of the performance of the TSHCT, the opponent team trained by the QMIX is used. Experiment results show that TSHCT outperforms the baseline algorithms after 60k episodes, as shown in Fig.~\ref{fig:rewards_vs_vdn}. The maximum average total rewards of TSHCT, QMIX, VDN, and COMA are 5.92, 4.98, 4.95, and 4.50, respectively.
The maximum averages of score-concede rate of TSHCT, QMIX, VDN, and COMA are 52.08\%, 46.63\%, 48.84\%, and 29.99\%, respectively, as shown in Fig.~\ref{fig:rate_vs_coma}. It is seen that the TSHCT improved the performance by 5.45\% as compared to that of QMIX. It is important to mention that QMIX is the algorithm with the best performance when compared with the other baseline methods, VDN and COMA.

The results presented in the team reward graphs in the comparisons with other CTDE algorithms, QMIX, VDN, and COMA, indicate that it is difficult to train for a cooperative behavior while performing multiple roles. The results of the proposed method suggest that, to solve this problem, additional optimization processes are required to take individual role rewards and shared team rewards into consideration. The proposed method uses training stage 2 to induce and learn a cooperative policy. From the observation that the role rewards of the TSHCT are greater than those of other methods, as shown in reward plots in Fig.~\ref{fig:rewards_vs_coma}, Fig.~\ref{fig:rewards_vs_vdn}, and Fig.~\ref{fig:rewards_vs_qmix}, it can be stated that the proposed method outperforms other baseline methods.
As shown in Fig.~\ref{fig:rate_vs_coma}, Fig.~\ref{fig:rate_vs_vdn}, and Fig.~\ref{fig:rate_vs_qmix}, the proposed method is able to achieve substantially higher score-concede rate than other methods. These results show that the proposed method can achieve improved performance against different opponent teams in competitive scenarios while learning policies.

\begin{table}[h!]
\centering
\begin{tabular}{|| c | c || c | c | c || }
 \hline
 \multicolumn{2}{||c||}{\textbf{TSHCT}} & vs \textbf{QMIX} & vs \textbf{VDN} & vs \textbf{COMA} \\ 
 \hline
 \hline
 \multirow{5}{*}{\textbf{\shortstack{100k \\ Episodes \\ Trained \\ Policy}}} & Score & 3.82 $\pm$ 1.70 & 3.92 $\pm$ 1.38 & 7.09 $\pm$ 1.83 \\  
 & Concede & 4.55 $\pm$ 0.89 & 4.23 $\pm$ 2.04 & 3.27 $\pm$ 1.54 \\  
 & Score Difference & -0.73 & -0.30 & 3.82 \\  
 & Score Concede Rate & 45.6\% & 48.1\% & 68.4\% \\  
 & Winning Rate & 20\% & 50\% & 100\% \\
 \hline
 \multirow{5}{*}{\textbf{\shortstack{200k \\ Episodes \\ Trained \\ Policy}}} & Score & 3.45 $\pm$ 1.44 & 5.00 $\pm$ 1.13 & 5.55 $\pm$ 2.23 \\  
 & Concede & 3.00 $\pm$ 1.41 & 2.82 $\pm$ 1.70 & 2.18 $\pm$ 1.59 \\  
 & Score Difference & 0.45 & 2.18 & 3.37 \\  
 & Score Concede Rate & 53.5\% & 63.9\% & 71.8\% \\  
 & Winning Rate & 80\% & 90\% & 100\% \\
 \hline
\end{tabular}
\caption{\label{tab:match_result}Results and statistics of evaluation matches for TSHCT against the baseline methods.}
\end{table}

The final performances of the policies trained by the proposed method and the baseline methods are compared by conducting 10 minutes matches. Table~\ref{tab:match_result} summarizes the results and statistics of these matches.

\subsubsection*{Ablation Study: DRQN vs Dueling DRQN}

In AI robot soccer, several different sequences of actions can lead to similar reward values. From this observation, an ablation study is conducted by combining the TSHCT with dueling Q-network. Dueling Q-network often leads to better policy in the presence of distinct actions leading to similar reward values \citep{wang2016dueling}. In this ablation study, the traditional dueling Q-network architecture is combined with the RNN, which is named here as Dueling DRQN. The proposed method combined with the Dueling DRQN is compared with the TSHCT combined with the DRQN. The TSHCT with Dueling DRQN is trained with 200k episodes using epsilon greedy exploration with self-play, similar to the cases shown in Fig.~\ref{fig:rate_vs_coma}, Fig.~\ref{fig:rate_vs_vdn}, and Fig.~\ref{fig:rate_vs_qmix}. For comparisons of rewards and score-concede rates, game matches between the team trained by the TSHCT with DRQN and the team trained by the TSHCT with Dueling DRQN are conducted. The results of these matches are presented in Table 2.

\begin{figure}[h!]
\centering
\includegraphics[width=0.85\linewidth]{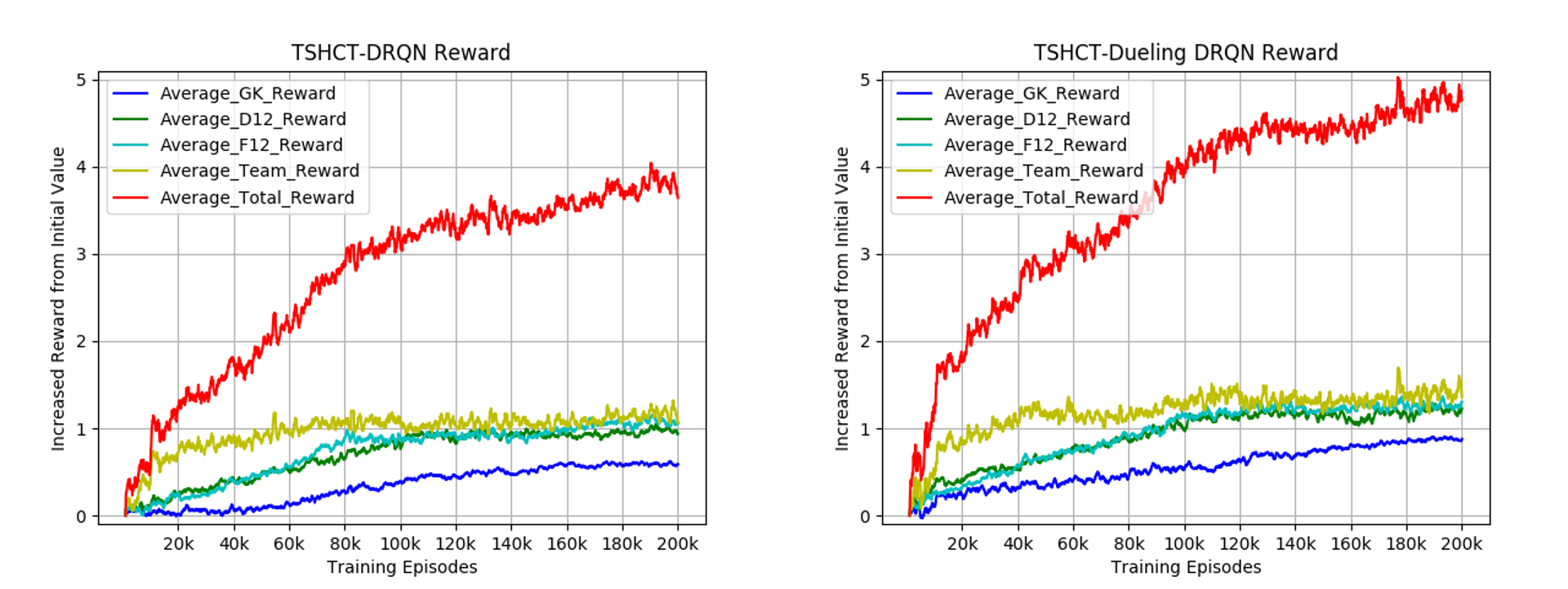}
\caption{Rewards of TSHCT with DRQN and Dueling DRQN during training for 200k episodes with self-play.}
\label{fig:reward_tshct}
\end{figure}

\begin{figure}[h!]
\centering
\includegraphics[width=\linewidth]{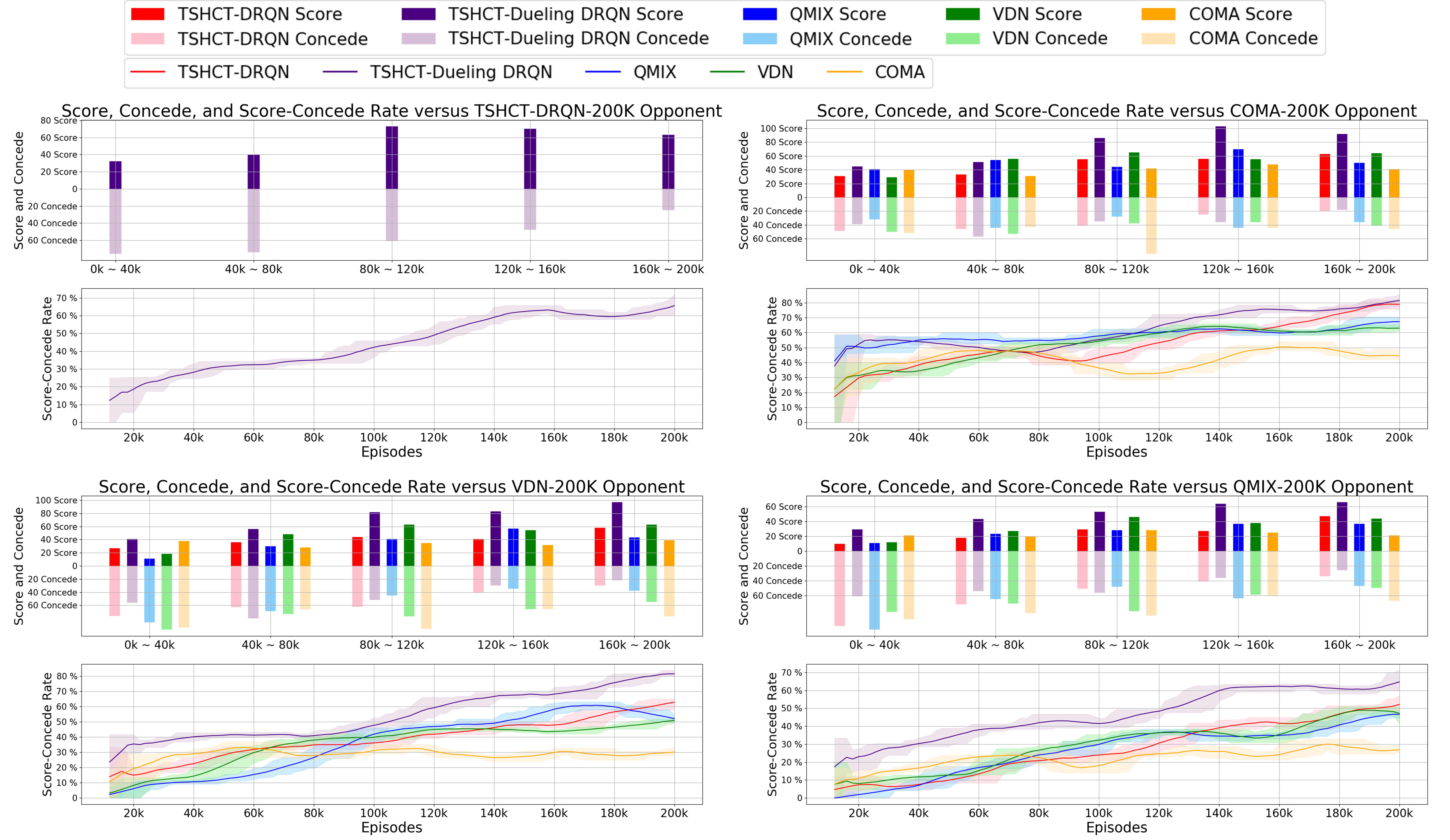}
\caption{Comparison of score, concede, and score-concede rate obtained during training by TSHCT with DRQN and Dueling DRQN, QMIX, VDN, and COMA evaluated against an opponent trained with baseline algorithms for 200k episodes.}
\label{fig:rate_dueling_vs_all}
\end{figure}

In Fig.~\ref{fig:reward_tshct}, the rewards obtained by the TSHCT with DRQN and the TSHCT with Dueling DRQN are presented. Figure 10 shows the increasing trends of rewards. It is seen that the TSHCT with Dueling DRQN leads to a higher total reward as compared to the TSHCT with DRQN.
The maximum average score-concede rates of the team trained by the TSHCT with Dueling DRQN with opponent teams trained by the TSHCT with DRQN, COMA, VDN, and QMIX are 65.59\%, 81.49\%, 81.52\%, and 64.67\%, respectively, as shown in Fig.~\ref{fig:rate_dueling_vs_all}. The TSHCT with Dueling DRQN demonstrates improved score-concede rates over COMA, VDN, and QMIX by 2.48\%, 18.67\%, and 12.59\% as compared to that obtained by the TSHCT with DRQN.

The policies trained by the TSHCT combined with DRQN and by the TSHCT combined with Dueling DRQN are compared with game results. Table~\ref{tab:match_result_drqn_vs_ddrqn} lists the results of these evaluation matches. For policies trained with the same number of training episodes, the TSHCT combined with Dueling DRQN outperforms the TSHCT combined with DRQN, achieving 60\% and 80\% winning rates with 100k episodes and 200k episodes, respectively. For the cases in which one algorithm is trained with two times the number of episodes of the opponent, i.e. 200k versus 100k, the algorithm that was trained for a longer time achieves a higher winning rate. However, even for this case, the trained policy using the TSHCT combined with Dueling DRQN is more robust, achieving a 30\% winning rate and a score-concede rate of 45.7\%.

\begin{table}[h!]
\centering
\begin{tabular}{|| c | c || c | c || }
 \hline
 \multicolumn{2}{||c||}{\multirow{3}{*}{\textbf{TSHCT-Dueling DRQN}}} & \multicolumn{2}{ c||}{vs \textbf{TSHCT-DRQN}} \\
 \cline{3-4}
 \multicolumn{2}{||c||}{} & \textbf{100k Episodes} & \textbf{200k Episodes} \\
 \multicolumn{2}{||c||}{} & \textbf{Trained Policy} & \textbf{Trained Policy} \\
 \hline
 \hline
 \multirow{5}{*}{\textbf{\shortstack{100k \\ Episodes \\ Trained \\ Policy}}} & Score & 5.09 $\pm$ 2.07 & 3.91 $\pm$ 1.78 \\  
 & Concede & 3.82 $\pm$ 2.03 & 4.64 $\pm$ 2.19 \\  
 & Score Difference & 1.27 & -0.73 \\  
 & Score Concede Rate & 57.1\% & 45.7\% \\  
 & Winning Rate & 60\% & 30\% \\
 \hline
 \multirow{5}{*}{\textbf{\shortstack{200k \\ Episodes \\ Trained \\ Policy}}} & Score & 8.36 $\pm$ 2.64 & 5.82 $\pm$ 2.48 \\  
 & Concede & 1.55 $\pm$ 1.30 & 2.27 $\pm$ 1.14  \\  
 & Score Difference & 6.81 & 3.55 \\  
 & Score Concede Rate & 84.4\% & 71.9\% \\  
 & Winning Rate & 100\% & 80\% \\
 \hline
\end{tabular}
\caption{\label{tab:match_result_drqn_vs_ddrqn}Results and statistics of evaluation matches for TSHCT-Dueling DRQN against TSHCT-DRQN.}
\end{table}

% === V. Conclusion ===
\section*{Conclusion}
\label{sec:5}

This paper deals with multi-agent reinforcement learning with heterogeneous agents. The classic way to solve this problem is using the CTDE framework. However, the CTDE framework is less efficient for heterogeneous agents in learning individual behaviors. This paper presents the TSHCT, a novel heterogeneous multi-agent reinforcement learning method that allows heterogeneous agents to learn multiple roles for cooperative behavior. In the proposed method, there are two training stages that are conducted in a serial manner. The first stage is for training individual behavior through maximizing individual role rewards, while the second stage is for training cooperative behavior while maximizing a shared team reward. The experiments are conducted with 5 versus 5 AI robot soccer which is relevant to the cooperative-competitive multi-agent environment. The proposed method is compared with other baseline methods that maximize the shared reward to achieve cooperative behavior. The proposed method and baseline methods, COMA, VDN, and QMIX, are combined with value-based algorithms, such as DQN and dueling Q-networks.

Comparisons of total rewards and score-concede rates are presented in the paper. The results show that the TSHCT training method is superior to other baseline algorithms in role training and learning cooperative behavior. The maximum average score-concede rates of the TSHCT in comparison with the COMA, VDN, and QMIX are 79.01\%, 62.85\%, and 52.08\%, respectively, representing the improvement achieved by the TSHCT in competitive AI robot soccer matches.

Because similar action-observation history leads to similar rewards in AI robot soccer, the training process can be unstable. To address this issue, an ablation study comparing the TSHCT combined with Dueling DRQN and DRQN is conducted. The performances of the TSHCT with DRQN and Dueling DRQN are measured by total rewards, score-concede rates, and match results. As a result, the TSHCT combined with Dueling DRQN achieves better performance when compared to the TSHCT combined with DRQN. The maximum average score-concede rate of the TSHCT with Dueling DRQN in comparison with the COMA, VDN, and QMIX are 81.49\%, 81.52\%, and 64.67\%, respectively. This result represents an improvement of 2.48\%, 18.67\%, and 12.59\% as compared to the case of the TSHCT combined with DRQN.

Simulation results show that the TSHCT is able to train an AI robot soccer team effectively, achieving higher individual role rewards and higher total rewards, as compared to other approaches that can be used for training to get cooperative behavior in a multi-agent environment. As future work, this framework is to be combined with actor-critic policy-based multi-agent algorithms that can be applied in environments with continuous actions.

\bibliography{references}

% Generated by IEEEtranN.bst, version: 1.14 (2015/08/26)
\begin{thebibliography}{37}
\providecommand{\natexlab}[1]{#1}
\providecommand{\url}[1]{#1}
\csname url@samestyle\endcsname
\providecommand{\newblock}{\relax}
\providecommand{\bibinfo}[2]{#2}
\providecommand{\BIBentrySTDinterwordspacing}{\spaceskip=0pt\relax}
\providecommand{\BIBentryALTinterwordstretchfactor}{4}
\providecommand{\BIBentryALTinterwordspacing}{\spaceskip=\fontdimen2\font plus
\BIBentryALTinterwordstretchfactor\fontdimen3\font minus
  \fontdimen4\font\relax}
\providecommand{\BIBforeignlanguage}[2]{{%
\expandafter\ifx\csname l@#1\endcsname\relax
\typeout{** WARNING: IEEEtranN.bst: No hyphenation pattern has been}%
\typeout{** loaded for the language `#1'. Using the pattern for}%
\typeout{** the default language instead.}%
\else
\language=\csname l@#1\endcsname
\fi
#2}}
\providecommand{\BIBdecl}{\relax}
\BIBdecl

\bibitem[Silver et~al.(2018)Silver, Hubert, Schrittwieser, Antonoglou, Lai,
  Guez, Lanctot, Sifre, Kumaran, Graepel, et~al.]{silver2018general}
D.~Silver, T.~Hubert, J.~Schrittwieser, I.~Antonoglou, M.~Lai, A.~Guez,
  M.~Lanctot, L.~Sifre, D.~Kumaran, T.~Graepel \emph{et~al.}, ``A general
  reinforcement learning algorithm that masters chess, shogi, and go through
  self-play,'' \emph{Science}, vol. 362, no. 6419, pp. 1140--1144, 2018.

\bibitem[Mnih et~al.(2015)Mnih, Kavukcuoglu, Silver, Rusu, Veness, Bellemare,
  Graves, Riedmiller, Fidjeland, Ostrovski, et~al.]{mnih2015human}
V.~Mnih, K.~Kavukcuoglu, D.~Silver, A.~A. Rusu, J.~Veness, M.~G. Bellemare,
  A.~Graves, M.~Riedmiller, A.~K. Fidjeland, G.~Ostrovski \emph{et~al.},
  ``Human-level control through deep reinforcement learning,'' \emph{nature},
  vol. 518, no. 7540, pp. 529--533, 2015.

\bibitem[Mnih et~al.(2016)Mnih, Badia, Mirza, Graves, Lillicrap, Harley,
  Silver, and Kavukcuoglu]{mnih2016asynchronous}
V.~Mnih, A.~P. Badia, M.~Mirza, A.~Graves, T.~Lillicrap, T.~Harley, D.~Silver,
  and K.~Kavukcuoglu, ``Asynchronous methods for deep reinforcement learning,''
  in \emph{International conference on machine learning}.\hskip 1em plus 0.5em
  minus 0.4em\relax PMLR, 2016, pp. 1928--1937.

\bibitem[Silver et~al.(2016)Silver, Huang, Maddison, Guez, Sifre, Van
  Den~Driessche, Schrittwieser, Antonoglou, Panneershelvam, Lanctot,
  et~al.]{silver2016mastering}
D.~Silver, A.~Huang, C.~J. Maddison, A.~Guez, L.~Sifre, G.~Van Den~Driessche,
  J.~Schrittwieser, I.~Antonoglou, V.~Panneershelvam, M.~Lanctot \emph{et~al.},
  ``Mastering the game of go with deep neural networks and tree search,''
  \emph{nature}, vol. 529, no. 7587, pp. 484--489, 2016.

\bibitem[Andrychowicz et~al.(2017)Andrychowicz, Wolski, Ray, Schneider, Fong,
  Welinder, McGrew, Tobin, Abbeel, and Zaremba]{andrychowicz2017hindsight}
M.~Andrychowicz, F.~Wolski, A.~Ray, J.~Schneider, R.~Fong, P.~Welinder,
  B.~McGrew, J.~Tobin, P.~Abbeel, and W.~Zaremba, ``Hindsight experience
  replay,'' \emph{arXiv preprint arXiv:1707.01495}, 2017.

\bibitem[Hwangbo et~al.(2019)Hwangbo, Lee, Dosovitskiy, Bellicoso, Tsounis,
  Koltun, and Hutter]{hwangbo2019learning}
J.~Hwangbo, J.~Lee, A.~Dosovitskiy, D.~Bellicoso, V.~Tsounis, V.~Koltun, and
  M.~Hutter, ``Learning agile and dynamic motor skills for legged robots,''
  \emph{Science Robotics}, vol.~4, no.~26, 2019.

\bibitem[Seo et~al.(2019)Seo, Vecchietti, Lee, and Har]{seo2019rewards}
M.~Seo, L.~F. Vecchietti, S.~Lee, and D.~Har, ``Rewards prediction-based credit
  assignment for reinforcement learning with sparse binary rewards,''
  \emph{IEEE Access}, vol.~7, pp. 118\,776--118\,791, 2019.

\bibitem[Vecchietti et~al.(2020)Vecchietti, Kim, Choi, Hong, and
  Har]{vecchietti2020batch}
L.~F. Vecchietti, T.~Kim, K.~Choi, J.~Hong, and D.~Har, ``Batch prioritization
  in multigoal reinforcement learning,'' \emph{IEEE Access}, vol.~8, pp.
  137\,449--137\,461, 2020.

\bibitem[Vinyals et~al.(2019)Vinyals, Babuschkin, Czarnecki, Mathieu, Dudzik,
  Chung, Choi, Powell, Ewalds, Georgiev, et~al.]{vinyals2019grandmaster}
O.~Vinyals, I.~Babuschkin, W.~M. Czarnecki, M.~Mathieu, A.~Dudzik, J.~Chung,
  D.~H. Choi, R.~Powell, T.~Ewalds, P.~Georgiev \emph{et~al.}, ``Grandmaster
  level in starcraft ii using multi-agent reinforcement learning,''
  \emph{Nature}, vol. 575, no. 7782, pp. 350--354, 2019.

\bibitem[Berner et~al.(2019)Berner, Brockman, Chan, Cheung, D{\k{e}}biak,
  Dennison, Farhi, Fischer, Hashme, Hesse, et~al.]{berner2019dota}
C.~Berner, G.~Brockman, B.~Chan, V.~Cheung, P.~D{\k{e}}biak, C.~Dennison,
  D.~Farhi, Q.~Fischer, S.~Hashme, C.~Hesse \emph{et~al.}, ``Dota 2 with large
  scale deep reinforcement learning,'' \emph{arXiv preprint arXiv:1912.06680},
  2019.

\bibitem[Liu et~al.(2019)Liu, Lever, Merel, Tunyasuvunakool, Heess, and
  Graepel]{liu2019emergent}
S.~Liu, G.~Lever, J.~Merel, S.~Tunyasuvunakool, N.~Heess, and T.~Graepel,
  ``Emergent coordination through competition,'' \emph{arXiv preprint
  arXiv:1902.07151}, 2019.

\bibitem[He et~al.(2015)He, Dai, and Ning]{he2015improving}
X.~He, H.~Dai, and P.~Ning, ``Improving learning and adaptation in security
  games by exploiting information asymmetry,'' in \emph{2015 IEEE Conference on
  Computer Communications (INFOCOM)}.\hskip 1em plus 0.5em minus 0.4em\relax
  IEEE, 2015, pp. 1787--1795.

\bibitem[Klima et~al.(2016)Klima, Tuyls, and Oliehoek]{klima2016markov}
R.~Klima, K.~Tuyls, and F.~Oliehoek, ``Markov security games: Learning in
  spatial security problems,'' in \emph{NIPS Workshop on Learning, Inference
  and Control of Multi-Agent Systems}, 2016, pp. 1--8.

\bibitem[Chu et~al.(2019)Chu, Wang, Codec{\`a}, and Li]{chu2019multi}
T.~Chu, J.~Wang, L.~Codec{\`a}, and Z.~Li, ``Multi-agent deep reinforcement
  learning for large-scale traffic signal control,'' \emph{IEEE Transactions on
  Intelligent Transportation Systems}, vol.~21, no.~3, pp. 1086--1095, 2019.

\bibitem[Zhang et~al.(2019)Zhang, Feng, Liu, Ding, Zhu, Zhou, Zhang, Yu, Jin,
  and Li]{zhang2019cityflow}
H.~Zhang, S.~Feng, C.~Liu, Y.~Ding, Y.~Zhu, Z.~Zhou, W.~Zhang, Y.~Yu, H.~Jin,
  and Z.~Li, ``Cityflow: A multi-agent reinforcement learning environment for
  large scale city traffic scenario,'' in \emph{The World Wide Web Conference},
  2019, pp. 3620--3624.

\bibitem[Shalev-Shwartz et~al.(2016)Shalev-Shwartz, Shammah, and
  Shashua]{shalev2016safe}
S.~Shalev-Shwartz, S.~Shammah, and A.~Shashua, ``Safe, multi-agent,
  reinforcement learning for autonomous driving,'' \emph{arXiv preprint
  arXiv:1610.03295}, 2016.

\bibitem[Sallab et~al.(2017)Sallab, Abdou, Perot, and Yogamani]{sallab2017deep}
A.~E. Sallab, M.~Abdou, E.~Perot, and S.~Yogamani, ``Deep reinforcement
  learning framework for autonomous driving,'' \emph{Electronic Imaging}, vol.
  2017, no.~19, pp. 70--76, 2017.

\bibitem[Nguyen et~al.(2020)Nguyen, Nguyen, and Nahavandi]{nguyen2020deep}
T.~T. Nguyen, N.~D. Nguyen, and S.~Nahavandi, ``Deep reinforcement learning for
  multiagent systems: A review of challenges, solutions, and applications,''
  \emph{IEEE transactions on cybernetics}, vol.~50, no.~9, pp. 3826--3839,
  2020.

\bibitem[Lowe et~al.(2017)Lowe, Wu, Tamar, Harb, Abbeel, and
  Mordatch]{lowe2017multi}
R.~Lowe, Y.~Wu, A.~Tamar, J.~Harb, P.~Abbeel, and I.~Mordatch, ``Multi-agent
  actor-critic for mixed cooperative-competitive environments,'' \emph{arXiv
  preprint arXiv:1706.02275}, 2017.

\bibitem[Sunehag et~al.(2017)Sunehag, Lever, Gruslys, Czarnecki, Zambaldi,
  Jaderberg, Lanctot, Sonnerat, Leibo, Tuyls, et~al.]{sunehag2017value}
P.~Sunehag, G.~Lever, A.~Gruslys, W.~M. Czarnecki, V.~Zambaldi, M.~Jaderberg,
  M.~Lanctot, N.~Sonnerat, J.~Z. Leibo, K.~Tuyls \emph{et~al.},
  ``Value-decomposition networks for cooperative multi-agent learning,''
  \emph{arXiv preprint arXiv:1706.05296}, 2017.

\bibitem[Foerster et~al.(2018)Foerster, Farquhar, Afouras, Nardelli, and
  Whiteson]{foerster2018counterfactual}
J.~Foerster, G.~Farquhar, T.~Afouras, N.~Nardelli, and S.~Whiteson,
  ``Counterfactual multi-agent policy gradients,'' in \emph{Proceedings of the
  AAAI Conference on Artificial Intelligence}, vol.~32, no.~1, 2018.

\bibitem[Samvelyan et~al.(2019)Samvelyan, Rashid, De~Witt, Farquhar, Nardelli,
  Rudner, Hung, Torr, Foerster, and Whiteson]{samvelyan2019starcraft}
M.~Samvelyan, T.~Rashid, C.~S. De~Witt, G.~Farquhar, N.~Nardelli, T.~G. Rudner,
  C.-M. Hung, P.~H. Torr, J.~Foerster, and S.~Whiteson, ``The starcraft
  multi-agent challenge,'' \emph{arXiv preprint arXiv:1902.04043}, 2019.

\bibitem[Rashid et~al.(2020)Rashid, Samvelyan, De~Witt, Farquhar, Foerster, and
  Whiteson]{rashid2020monotonic}
T.~Rashid, M.~Samvelyan, C.~S. De~Witt, G.~Farquhar, J.~Foerster, and
  S.~Whiteson, ``Monotonic value function factorisation for deep multi-agent
  reinforcement learning,'' \emph{Journal of Machine Learning Research},
  vol.~21, no. 178, pp. 1--51, 2020.

\bibitem[Lillicrap et~al.(2015)Lillicrap, Hunt, Pritzel, Heess, Erez, Tassa,
  Silver, and Wierstra]{lillicrap2015continuous}
T.~P. Lillicrap, J.~J. Hunt, A.~Pritzel, N.~Heess, T.~Erez, Y.~Tassa,
  D.~Silver, and D.~Wierstra, ``Continuous control with deep reinforcement
  learning,'' \emph{arXiv preprint arXiv:1509.02971}, 2015.

\bibitem[Mnih et~al.(2013)Mnih, Kavukcuoglu, Silver, Graves, Antonoglou,
  Wierstra, and Riedmiller]{mnih2013playing}
V.~Mnih, K.~Kavukcuoglu, D.~Silver, A.~Graves, I.~Antonoglou, D.~Wierstra, and
  M.~Riedmiller, ``Playing atari with deep reinforcement learning,''
  \emph{arXiv preprint arXiv:1312.5602}, 2013.

\bibitem[Hausknecht and Stone(2015)]{hausknecht2015deep}
M.~Hausknecht and P.~Stone, ``Deep recurrent q-learning for partially
  observable mdps,'' \emph{arXiv preprint arXiv:1507.06527}, 2015.

\bibitem[Wang et~al.(2016)Wang, Schaul, Hessel, Hasselt, Lanctot, and
  Freitas]{wang2016dueling}
Z.~Wang, T.~Schaul, M.~Hessel, H.~Hasselt, M.~Lanctot, and N.~Freitas,
  ``Dueling network architectures for deep reinforcement learning,'' in
  \emph{International conference on machine learning}.\hskip 1em plus 0.5em
  minus 0.4em\relax PMLR, 2016, pp. 1995--2003.

\bibitem[{Hong} et~al.(2021){Hong}, {Jeong}, {Vecchietti}, {Har}, and
  {Kim}]{hong2021}
C.~{Hong}, I.~{Jeong}, L.~F. {Vecchietti}, D.~{Har}, and J.~H. {Kim}, ``Ai
  world cup: Robot soccer-based competitions,'' \emph{IEEE Transactions on
  Games}, pp. 1--1, 2021.

\bibitem[Heinrich et~al.(2015)Heinrich, Lanctot, and
  Silver]{heinrich2015fictitious}
J.~Heinrich, M.~Lanctot, and D.~Silver, ``Fictitious self-play in
  extensive-form games,'' in \emph{International conference on machine
  learning}.\hskip 1em plus 0.5em minus 0.4em\relax PMLR, 2015, pp. 805--813.

\bibitem[Lanctot et~al.(2017)Lanctot, Zambaldi, Gruslys, Lazaridou, Tuyls,
  P{\'e}rolat, Silver, and Graepel]{lanctot2017unified}
M.~Lanctot, V.~Zambaldi, A.~Gruslys, A.~Lazaridou, K.~Tuyls, J.~P{\'e}rolat,
  D.~Silver, and T.~Graepel, ``A unified game-theoretic approach to multiagent
  reinforcement learning,'' \emph{arXiv preprint arXiv:1711.00832}, 2017.

\bibitem[Silver et~al.(2017)Silver, Schrittwieser, Simonyan, Antonoglou, Huang,
  Guez, Hubert, Baker, Lai, Bolton, et~al.]{silver2017mastering}
D.~Silver, J.~Schrittwieser, K.~Simonyan, I.~Antonoglou, A.~Huang, A.~Guez,
  T.~Hubert, L.~Baker, M.~Lai, A.~Bolton \emph{et~al.}, ``Mastering the game of
  go without human knowledge,'' \emph{nature}, vol. 550, no. 7676, pp.
  354--359, 2017.

\bibitem[Oliehoek and Amato(2016)]{oliehoek2016concise}
F.~A. Oliehoek and C.~Amato, \emph{A concise introduction to decentralized
  POMDPs}.\hskip 1em plus 0.5em minus 0.4em\relax Springer, 2016.

\bibitem[Hochreiter and Schmidhuber(1997)]{hochreiter1997long}
S.~Hochreiter and J.~Schmidhuber, ``Long short-term memory,'' \emph{Neural
  computation}, vol.~9, no.~8, pp. 1735--1780, 1997.

\bibitem[Chung et~al.(2014)Chung, Gulcehre, Cho, and
  Bengio]{chung2014empirical}
J.~Chung, C.~Gulcehre, K.~Cho, and Y.~Bengio, ``Empirical evaluation of gated
  recurrent neural networks on sequence modeling,'' \emph{arXiv preprint
  arXiv:1412.3555}, 2014.

\bibitem[Ha et~al.(2016)Ha, Dai, and Le]{ha2016hypernetworks}
D.~Ha, A.~Dai, and Q.~V. Le, ``Hypernetworks,'' \emph{arXiv preprint
  arXiv:1609.09106}, 2016.

\bibitem[Michel(2004)]{michel2004cyberbotics}
O.~Michel, ``Cyberbotics ltd. webots™: professional mobile robot
  simulation,'' \emph{International Journal of Advanced Robotic Systems},
  vol.~1, no.~1, p.~5, 2004.

\bibitem[Kingma and Ba(2014)]{kingma2014adam}
D.~P. Kingma and J.~Ba, ``Adam: A method for stochastic optimization,''
  \emph{arXiv preprint arXiv:1412.6980}, 2014.

\end{thebibliography}

\end{document}